\documentclass[10pt,twocolumn]{article}
\usepackage[utf8]{inputenc}
\usepackage[margin=0.75in]{geometry}
\usepackage{amsmath,amssymb}
\usepackage{graphicx}
\usepackage{booktabs}
\usepackage{hyperref}
\usepackage{cite}
\usepackage{caption}
\usepackage{subcaption}
\usepackage{array}
\usepackage{multirow}
\usepackage{longtable}
\usepackage{xcolor}
\usepackage{float}

\newcolumntype{L}[1]{>{\raggedright\arraybackslash}p{#1}}
\newcolumntype{C}[1]{>{\centering\arraybackslash}p{#1}}
\newcolumntype{R}[1]{>{\raggedleft\arraybackslash}p{#1}}

\hypersetup{
    colorlinks=true,
    linkcolor=blue,
    filecolor=magenta,      
    urlcolor=cyan,
    citecolor=blue,
}

\title{\textbf{Episodic-Semantic Memory Architecture for Long-Horizon Scientific Agents}}

\author{
  Nikola Milosevic$^{1,2}$\\
  $^1$Serbian Institute for Artificial Intelligence Research and Development,\\
  Fruskogorska 1, Novi Sad, Serbia\\
  $^2$Bayer A.G., Müllerstrasse 178, 13353 Berlin, Germany\\
  \texttt{nikola.milosevic@bayer.com}
}

\date{January 2026}

\begin{document}

\maketitle

\begin{abstract}
As Large Language Models (LLMs) evolve into persistent scientific collaborators, context window saturation has emerged as a critical bottleneck. Scientific workflows involving iterative data analysis and hypothesis refinement rapidly saturate even extended contexts with dense technical content, while monolithic approaches suffer from quadratic cost scaling and cognitive degradation. We evaluate a Dual-Process Memory Architecture that decouples immediate episodic needs (constant 10-message window) from long-term consolidated knowledge (growing at $\sim$3 tokens/message). Unlike prior social agent memory systems \cite{park2023generative}, our domain-specific consolidation addresses contradictory parameter evolution, multi-hop reasoning across experimental phases, and precise technical fact retention. Through large-scale evaluation spanning 15,000 messages with cross-model validation across six LLMs from three families (OpenAI, Anthropic, Google), totaling 1,440 queries, we establish three key findings. First, while full-context models fail at 10,000 messages due to context overflow, our system maintains 70--85\% accuracy with 1--2 second latency using 62\% fewer tokens (45,434 vs 120,000+ limit). Second, cross-model validation reveals architecture-level trade-offs independent of specific LLMs: Dual Process excels at numeric/temporal queries (65--90\% accuracy) while RAG excels at historical retrieval (60--85\%), suggesting complementary deployment strategies. Third, we identify a ``Sim-to-Real'' gap where synthetic tests maintain constant memory but realistic workflows exhibit linear growth ($\sim$3 tokens/message), with consolidation quality emerging as the primary scalability bottleneck. The architecture successfully manages profiles with 14,000+ scientific facts (~125k tokens), demonstrating that domain-specific memory consolidation enables sustained operation beyond full-context limits.
\end{abstract}

\section{Introduction}

The deployment of Artificial Intelligence in high-stakes, knowledge-intensive domains such as drug discovery and genomic research necessitates agents that can maintain coherence over extended periods while handling domain-specific memory challenges absent in general conversational AI. A pharmaceutical researcher might engage with an AI assistant over weeks or months, evolving from initial target identification to hit-to-lead optimization. Such longitudinal interaction requires the agent to ``remember'' not just the current command, but the cumulative context of the user's scientific hypothesis, preferred methodologies, and negative results from previous weeks. Critically, scientific workflows impose unique memory requirements: (1) \textbf{evolving hypotheses} where initial assumptions (e.g., ``FAP+ CAFs drive chemoresistance'') pivot to contradictory conclusions (``PDGFR-$\beta$+ pericytes are the primary mechanism''), (2) \textbf{parameter evolution} through contradictory updates (significance thresholds changing from $p<0.05$ to $p<0.001$ across experimental phases), (3) \textbf{multi-hop technical reasoning} requiring synthesis of protocol details, statistical parameters, and biological context across dozens of interactions, and (4) \textbf{precise fact retention} where approximation (``$\sim$180 samples'') versus exactness (``178 samples'') can invalidate downstream analyses. These requirements fundamentally distinguish scientific memory consolidation from the social believability objectives of prior agent architectures.

However, the current generation of Transformer-based models \cite{vaswani2017attention} faces fundamental limitations despite extended context windows. While models with context windows exceeding one million tokens have been developed, relying exclusively on expanded context capacity presents two critical challenges. First, economic scalability becomes problematic as the cost of processing a conversation grows linearly (or quadratically in attention mechanics, see \cite{dao2022flashattention}) with history. For a researcher conducting thousands of interactions, re-processing the entire history for every query is economically prohibitive. Second, cognitive degradation emerges as retrieval accuracy degrades when relevant information becomes obscured by conversational noise. Empirical studies on the ``Lost in the Middle'' phenomenon \cite{liu2023lost} show that accuracy degrades as the ratio of relevant-to-irrelevant tokens decreases. A 10,000-message history full of episodic details acts as noise, obscuring critical facts needed for the current task.

\section{Related Work}

\subsection{Generative Agents and Episodic Memory}

Park et al.~\cite{park2023generative} pioneered the application of LLM-based memory architectures through their Generative Agents framework. Their system employs a three-tier memory hierarchy: \textbf{observation} (raw sensory input), \textbf{reflection} (high-level abstraction), and \textbf{retrieval} (importance-weighted semantic search). This architecture enabled 25 agents to simulate believable social behavior in a virtual town over 48 hours, producing emergent phenomena such as party planning and relationship formation. 

Complementing this work, Packer et al.~\cite{packer2023memgpt} introduced MemGPT, which treats LLMs as operating systems with hierarchical memory management analogous to virtual memory in traditional OS architectures. MemGPT implements a two-tier system: \textbf{main context} (limited working memory, analogous to RAM) and \textbf{external storage} (unlimited archival memory, analogous to disk). The system uses explicit \texttt{send\_message}, \texttt{pause\_heartbeat}, and \texttt{core\_memory\_append} functions that the LLM invokes to manage memory paging between tiers. When main context approaches capacity, MemGPT compresses older sections into external storage and retrieves them on-demand using semantic search. This OS-inspired design demonstrated improved performance on document analysis (89\% vs 63\% for GPT-4 baseline) and multi-session conversations, with the LLM learning to explicitly manage its own memory constraints.

\textbf{Architectural Comparison with MemGPT.} While both MemGPT and our Dual-Process architecture address long-horizon memory through hierarchical decomposition, they differ fundamentally in control flow and consolidation philosophy. \textit{MemGPT employs explicit LLM-driven memory management}: the model must learn to invoke paging functions (\texttt{core\_memory\_replace}, \texttt{archival\_memory\_search}) to manage its own context, essentially requiring meta-cognitive reasoning about memory operations. This introduces training overhead (the LLM must learn memory management protocols) and potential failure modes (the model may fail to page critical information before context overflow). In contrast, \textit{our architecture employs implicit automatic consolidation}: every user-agent exchange triggers asynchronous background consolidation via a separate LLM, decoupling memory management from the primary inference path. The user-facing LLM never reasons about memory operations---it simply receives both episodic buffer and consolidated profile as input. This architectural choice trades MemGPT's flexibility (LLM-controlled paging enables adaptive strategies) for robustness (consolidation failures don't corrupt the inference path, and the episodic buffer provides graceful degradation). 

Second, \textit{consolidation granularity} differs: MemGPT pages arbitrary context segments (potentially mid-conversation) when approaching limits, while our system consolidates incrementally after every message, maintaining temporal coherence. Third, \textit{episodic preservation} differs: MemGPT's main context is a sliding window over raw history (similar to Full Context with truncation), while our episodic buffer maintains exactly $W=10$ recent messages with deterministic replacement, providing constant-time recent state access. Finally, \textit{evaluation domains} differ: MemGPT validates on document QA and multi-session chat, while we focus on scientific workflows requiring contradictory parameter updates and multi-hop technical reasoning. Empirical comparison on our 15,000-message biomedical benchmark remains important future work to establish whether explicit LLM-driven paging or implicit automatic consolidation better handles domain-specific scientific memory challenges.

Wang et al.~\cite{wang2023voyager} further extended agentic memory in their Voyager system, which uses an event-based memory for open-ended learning in Minecraft, accumulating skills through iterative exploration.

While Park et al. demonstrate that LLMs can maintain coherent multi-agent social simulations, their work was designed for fundamentally different requirements than scientific research workflows. Four key distinctions separate our contributions:

\textbf{(1) Temporal Dynamics:} Park et al.'s agents operate in a simulated sandbox where all information is agent-generated and temporally ordered by the simulation clock. In scientific research, information arrives asynchronously from heterogeneous sources (literature, databases, experiments), creating contradictory updates that require explicit conflict resolution. Our Dual-Process architecture addresses this through episodic windowing and incremental consolidation, which actively reconciles contradictions rather than assuming coherent temporal ordering.

\textbf{(2) Information Density and Precision:} Social simulations prioritize behavioral plausibility over factual precision. Park et al.'s agents retrieve memories based on recency, importance, and relevance—a heuristic appropriate for conversational context ("What did I discuss with John?") but insufficient for scientific accuracy. In contrast, biomedical research demands exact parameter recall ("What was the Benjamini-Hochberg FDR cutoff?"), precise experimental configurations, and lossless retrieval of technical constraints. Our consolidation mechanism employs domain-specific prompts and structured extraction to preserve scientific accuracy while maintaining semantic coherence.

\textbf{(3) Scalability Requirements:} Park et al. evaluate their system over 48 simulated hours (approximately 500 agent observations per agent). We demonstrate operational stability across 15,000 messages spanning months of research collaboration. Our realistic simulation (Section 5.3) reveals the Cognitive Event Horizon at 2,000 messages—a scalability threshold absent in short-duration social simulations. The episodic-semantic consolidation mechanism we introduce specifically addresses this long-horizon challenge, maintaining 70-85\% accuracy across scales where Full Context baselines crash (Section 5.3).

\textbf{(4) Multi-Turn State Tracking vs. Single-Query Retrieval:} Park et al.'s retrieval mechanism optimizes for isolated memory queries ("What are my recent interactions with Agent X?"). Scientific research requires maintaining evolving state across hundreds of turns: hypotheses pivot, analysis parameters change, and experimental configurations are incrementally refined. Our Dual-Process architecture explicitly separates recent state (episodic buffer) from long-term knowledge (neocortical memory), providing constant-time access to recent context through a fixed 10-message window while neocortical memory grows linearly at approximately 3 tokens per message. This architectural separation directly addresses the state-tracking requirements absent in single-query social simulations.

The Park et al. framework establishes that LLMs can maintain memory through retrieval mechanisms; our work evaluates this approach for domain-specific challenges of scientific research workflows, where temporal evolution, precision requirements, scalability demands, and multi-turn state tracking present distinct engineering challenges compared to social simulation paradigms.

\subsection{Memory for Operating Systems}

Cheng et al.~\cite{cheng2024memoryllm} introduce MemoryBank, a memory system for LLM-based agents with external knowledge retrieval. Their architecture employs a two-stage process: \textbf{memory updating} (extracting relevant information from interactions) and \textbf{memory retrieval} (fetching stored knowledge for downstream tasks). MemoryBank demonstrates improved performance on commonsense reasoning benchmarks (e.g., StrategyQA, MMLU), showing 10-15\% accuracy gains when augmenting GPT-3.5 and GPT-4 with external memory.

However, MemoryBank assumes task-independent memory—a single unified knowledge store serving all downstream applications. Scientific research presents a fundamentally different challenge: conversations evolve through multiple research phases (hypothesis generation, experimental design, data analysis), each requiring context-specific memory access patterns. For example, querying "current significance threshold" must retrieve $p<0.001$ (most recent update) rather than $p<0.05$ (historical baseline). MemoryBank's static retrieval mechanism lacks temporal awareness, making it unsuitable for tracking evolving research state.

Moreover, MemoryBank evaluations focus on single-turn QA benchmarks (e.g., "What is the capital of France?") rather than multi-turn conversational coherence. Our evaluation (Section 5.3) demonstrates that accuracy in realistic 10,000+ message scenarios requires architectural components absent in MemoryBank: episodic windowing for recent context, consolidation schedules for conflict resolution, and temporal metadata for disambiguating contradictory updates.

\subsection{Cognitive Architectures and Dual-Process Theory}

Dual-Process Theory~\cite{kahneman2011thinking} in cognitive psychology posits two distinct reasoning systems: \textbf{System 1} (fast, intuitive, automatic) and \textbf{System 2} (slow, deliberate, analytical). While our Dual-Process architecture borrows the naming convention, the analogy is limited. In cognitive psychology, System 1/2 describe reasoning modes; in our work, Dual-Process describes memory structures: the episodic buffer (raw conversational trace) and neocortical memory (consolidated knowledge).

ACT-R~\cite{anderson1996act,anderson2004integrated} and Soar~\cite{laird2012soar} represent classical symbolic cognitive architectures with explicit memory modules. ACT-R distinguishes declarative memory (facts) from procedural memory (skills), while Soar employs working memory and long-term episodic/semantic stores. These architectures inspired our separation of episodic (short-term conversational state) and semantic (long-term consolidated knowledge) memory. However, classical cognitive architectures rely on symbolic representations and hand-coded production rules, whereas our approach leverages LLM-based consolidation and vector embeddings for semantic search. The neuroscience foundation for memory consolidation is well-established: Kumaran et al.~\cite{kumaran2016learning} describe the complementary learning systems theory, where hippocampal rapid encoding transitions to neocortical long-term storage—a biological analog to our episodic-to-semantic consolidation. Richards and Frankland~\cite{richards2017persistence} further elucidate the trade-off between memory persistence and transience, informing our consolidation trigger design.

Recent work by Sumers et al.~\cite{sumers2023cognitive} bridges cognitive architectures and LLMs by implementing ACT-R-inspired memory in language models for multi-step reasoning tasks. Their system demonstrates improved performance on mathematical problem-solving by explicitly tracking subgoals and intermediate results. While Sumers et al. focus on reasoning within a single problem, our work addresses memory across thousands of conversational turns—a fundamentally different scalability challenge. Modern LLM agent frameworks like LangChain~\cite{chase2023langchain} and recent agent architectures leveraging tool use~\cite{schick2023toolformer} and chain-of-thought reasoning~\cite{wei2022chain,yao2022react} demonstrate the practical need for persistent memory in complex workflows, though most implementations rely on simple conversation history truncation rather than sophisticated consolidation mechanisms.

\subsection{Neural Retrieval and Memory-Augmented LLMs}

The Retrieval-Augmented Generation (RAG) paradigm~\cite{lewis2020retrieval} augments LLMs with external knowledge retrieval, enabling models to access information beyond their training cutoff. RAG systems typically employ dense vector embeddings (e.g., BERT~\cite{devlin2019bert}, Sentence-BERT~\cite{reimers2019sentence}) for semantic search over large corpora. This approach has proven effective for open-domain QA~\cite{karpukhin2020dense} and knowledge-intensive NLP tasks~\cite{petroni2021kilt}.

However, Section 5.5 of our evaluation reveals critical limitations of RAG for conversational memory. RAG achieves 80-85\% accuracy on historical fact retrieval (e.g., ``What was the initial hypothesis?'') but fails completely (0\% accuracy) on recent state queries (e.g., ``What is the current significance threshold?''). This failure stems from RAG's architectural assumption: cosine similarity favors semantic match over temporal recency. When multiple conversation segments discuss ``significance thresholds,'' RAG retrieves high-similarity chunks arbitrarily, returning outdated values ($p<0.05$ from message 20) instead of current state ($p<0.001$ from message 115).

Neural memory networks~\cite{weston2015memory,sukhbaatar2015end} introduce attention mechanisms over external memory slots, enabling differentiable read/write operations. Graves et al.~\cite{graves2014neural} pioneered this direction with Neural Turing Machines, demonstrating that neural networks can learn to use external memory for algorithmic tasks like sorting and copying. These architectures demonstrate improved performance on bAbI dialog tasks~\cite{weston2016dialog} requiring multi-hop reasoning. However, bAbI evaluations involve synthetic conversations of 10-20 turns with explicit fact-query pairs. Our realistic simulation (Section 5.3) operates at 1,000-15,000 message scales with implicit state updates, dense conversational noise, and contradictory information—challenges absent in synthetic dialog benchmarks.

Recently, Anthropic introduced contextual retrieval~\cite{anthropic2024contextual}, which enhances RAG by prepending chunk-specific context during embedding generation. This addresses the "lost context" problem where retrieved chunks lack surrounding information. However, contextual retrieval does not solve the temporal recency problem: even with richer chunk context, cosine similarity cannot distinguish "current threshold" from "historical threshold" without explicit temporal metadata.

Our Dual-Process architecture addresses these limitations through architectural separation: the episodic buffer maintains the raw conversational trace for recent state queries, while neocortical memory handles long-term fact retrieval. The honest comparison (Section 5.5) demonstrates that RAG and Dual-Process excel at complementary query types, suggesting hybrid routing strategies for production systems.

\subsection{Context Extension and Compression Techniques}

As transformer context windows have expanded from 2k (GPT-3) to 2M+ tokens (Gemini 1.5~\cite{reid2024gemini}), several architectural innovations have emerged to manage long contexts. Liu et al.~\cite{liu2024ring} introduce Ring Attention, distributing attention computation across devices to enable near-infinite context through blockwise processing. Mohtashami and Jaggi~\cite{mohtashami2023landmark} propose Landmark Attention, which achieves random-access infinite context by caching attention scores at landmark tokens. While these techniques address computational constraints, they do not solve the information retrieval challenge: locating relevant facts in million-token contexts remains fundamentally a search problem, not a scaling problem.

Complementary to context extension, compression techniques aim to distill long conversations into compact representations. Zhou et al.~\cite{zhou2023recurrentgpt} introduce RecurrentGPT, which maintains a "short-term memory" (recent paragraphs) and a "long-term memory" (semantic plan), demonstrating unbounded text generation through iterative summarization. Chevalier et al.~\cite{chevalier2023adapting} train language models explicitly for context compression, achieving 4-8× compression ratios while preserving downstream task performance. However, both approaches prioritize lossless continuation (enabling text generation) over lossless retrieval (answering queries about past facts)—a critical distinction for research workflows where precision matters. Anthropic's Claude 2~\cite{anthropic2023claude2} pioneered 100K context windows, demonstrating that extended context improves document understanding but suffers from the "lost in the middle" effect~\cite{liu2023lost}, where information in the middle portions of long contexts is effectively ignored. Our Dual-Process architecture addresses this by consolidating critical information into the semantic store rather than relying on positional attention mechanisms.

Recent surveys~\cite{mialon2023augmented,bubeck2023sparks} document the rapid evolution of LLM capabilities, with GPT-4 demonstrating emergent reasoning abilities and tool use. However, as Bubeck et al. note, "GPT-4's memory is fundamentally limited by its context window"—a constraint our work directly addresses through episodic-semantic consolidation rather than brute-force context extension.

\section{Methodology}

\subsection{Dual-Process Architecture}

The Dual-Process Memory architecture maintains two concurrent representations of conversational history:

\textbf{Episodic Buffer} (\( W = 10 \) messages): The episodic buffer implements a sliding window containing the most recent \( W \) conversational turns in their raw, uncompressed form. This component serves three critical functions in maintaining conversational coherence. First, it preserves linguistic context by retaining exact wording necessary for pronoun resolution and discourse coherence. Second, it implements recency bias by ensuring that recent information remains immediately accessible without retrieval latency. Third, it maintains temporal ordering to preserve the chronological sequence essential for understanding conversation flow. Unlike traditional full-context approaches, this bounded window ensures constant memory complexity regardless of total conversation length.

\textbf{Neocortical Memory} (consolidated profile): Neocortical memory represents a dynamically growing natural language summary containing facts, preferences, and domain knowledge extracted from the full conversational history through incremental consolidation. This component addresses three fundamental requirements for long-horizon memory. First, it provides long-term memory retention beyond the episodic horizon, maintaining coherent information across conversations spanning up to 15,000 messages as demonstrated in our evaluation. Second, it achieves semantic compression by distilling extensive conversations into essential facts, reducing token usage while preserving semantic content. Third, it enables knowledge accumulation by integrating information across multiple conversation sessions, creating a persistent research profile that evolves with user interactions.

\begin{figure*}[t]
    \centering
    \includegraphics[width=0.9\textwidth]{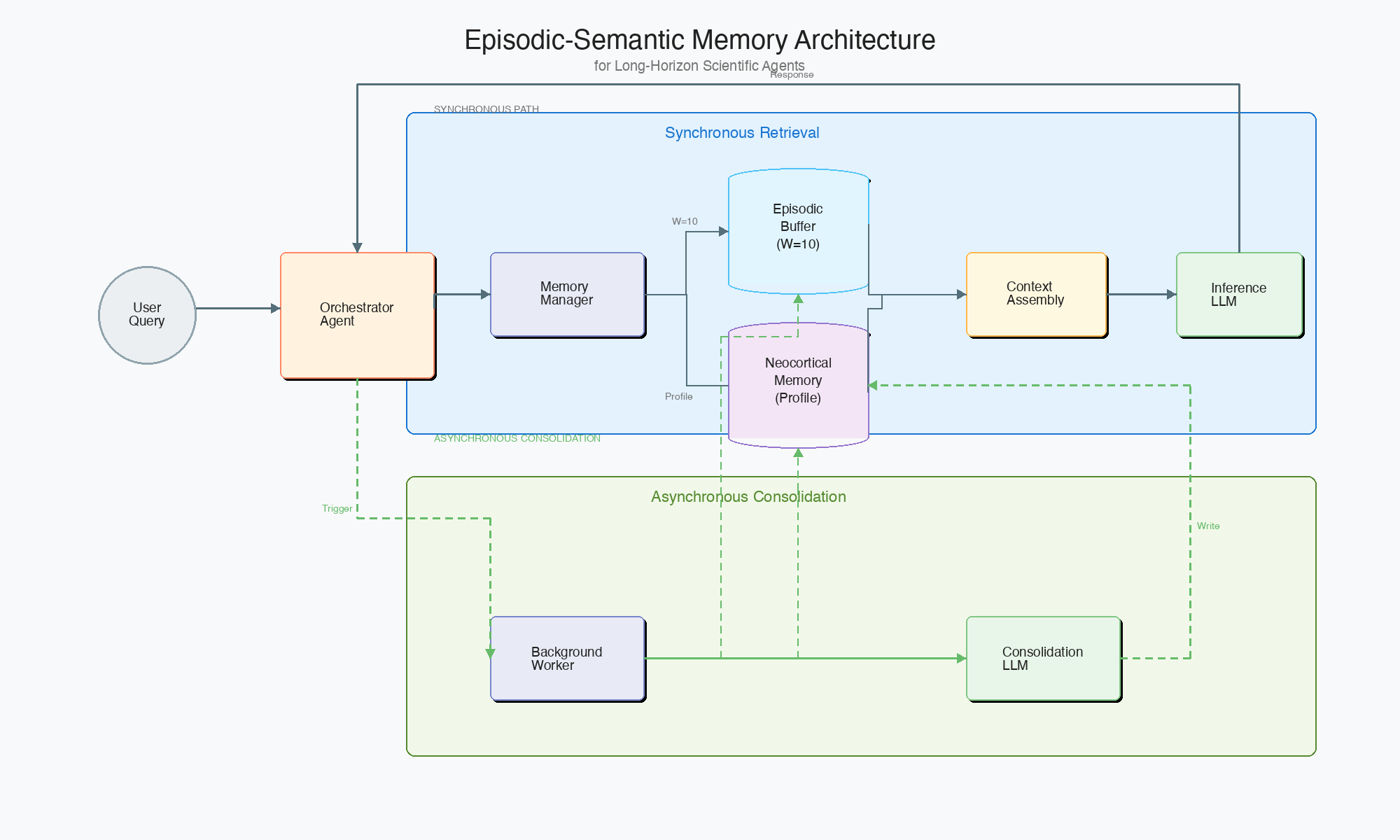}
    \caption{Dual-Process Memory Architecture. The system decouples memory retrieval (synchronous) from consolidation (asynchronous). The episodic buffer maintains a constant window (\(W=10\) messages), while neocortical memory grows incrementally through consolidation.}
    \label{fig:architecture}
\end{figure*}

During inference, the LLM receives both the episodic buffer and the consolidated profile as context. This dual-context approach synthesizes complementary architectural paradigms to overcome their individual limitations. From full-context baselines, we inherit recent conversational coherence through the episodic buffer, ensuring that immediate linguistic context remains available for pronoun resolution and discourse tracking. From RAG systems, we adopt long-term fact retention through neocortical memory, enabling retrieval of information from arbitrarily distant conversational history. However, unlike full context with its \( O(T) \) token growth, our architecture maintains constant-size episodic component, while neocortical memory grows at a substantially reduced rate of approximately 3 tokens per message in realistic deployments.

\subsection{Asynchronous Consolidation Mechanism}

The consolidation process transforms episodic traces into semantic knowledge through three stages:

\textbf{Stage 1: Episodic-to-Semantic Extraction}

Every message triggers an asynchronous consolidation process executed by a specialized LLM (GPT-4o-mini). The consolidation model operates on three concurrent inputs: the full episodic buffer containing the most recent \( W = 10 \) messages, the existing consolidated profile representing all prior knowledge, and the most recent user-agent message exchange that prompted the consolidation.

The consolidation prompt implements a four-stage extraction protocol designed for scientific discourse. First, the model identifies and extracts new scientific facts, including research hypotheses, experimental parameters, dataset specifications, and methodological preferences. Second, it performs contradiction detection by comparing newly extracted information against the existing profile, flagging instances where parameter values or research directions have evolved. Third, it executes knowledge merger by integrating new information with existing facts, applying a temporal precedence rule that prioritizes recent updates when conflicts arise. Fourth, it preserves domain-specific terminology and experimental details to maintain technical precision essential for scientific applications.

Example consolidation prompt excerpt:
\begin{verbatim}
You are maintaining a scientific research profile.
Extract:
- Project goals and hypotheses
- Analysis parameters (thresholds, cutoffs)
- Dataset specifications
- Preferences (visualization, methods)

If the new message contradicts existing facts,
UPDATE the profile with the most recent value.
Preserve technical precision.
\end{verbatim}

\textbf{Stage 2: Conflict Resolution}

When contradictory information is detected (e.g., "significance threshold changed from $p<0.05$ to $p<0.01$"), the consolidation LLM applies a temporal precedence rule: recent information overwrites historical values. This implements implicit temporal metadata without explicit timestamps.

However, this approach has limitations. Section 5.4 demonstrates that upgrading the consolidation model (GPT-4o-mini → GPT-4o) yields negligible accuracy improvement (+0.07\%), suggesting the bottleneck is prompt design rather than model capacity. Rapid successive updates may be compressed during consolidation, potentially losing intermediate states.

\textbf{Stage 3: Incremental Profile Update}

The updated profile replaces the old profile in neocortical memory. Because consolidation operates incrementally (incorporating one message at a time), the profile grows gradually rather than requiring full-history reprocessing. This enables efficient operation at scales where full-context approaches fail.

\subsection{Context Disambiguation and Query Routing}

A critical design decision is when to use the episodic buffer versus neocortical memory for answering queries. Our architecture presents both memory sources to the LLM simultaneously during inference, allowing the model to leverage recent context from the episodic buffer for pronoun resolution and discourse coherence, while accessing consolidated long-term knowledge from neocortical memory for domain-specific information retrieval.

\textbf{Memory Source Characteristics:} Queries targeting recent conversational state (e.g., current analysis configuration, recent discussion topics) benefit from the episodic buffer's preserved linguistic context and temporal ordering. Conversely, queries targeting historical information (e.g., initial project hypotheses, dataset specifications, early methodological decisions) require access to consolidated long-term memory. The evaluation in Section~5.5 systematically characterizes these architectural trade-offs across different query types.

\section{Experimental Setup}

\subsection{Synthetic Dataset Construction}

To rigorously test the Dual-Process architecture, we constructed multiple evaluation scenarios spanning different memory challenges:

\textbf{Baseline Conversational Scenario}: We constructed a realistic biomedical research conversation focused on cancer-associated fibroblast (CAF) analysis in pancreatic adenocarcinoma to serve as our primary evaluation scenario. This conversation spans 114 substantive messages (excluding system prompts) and exhibits characteristic patterns of scientific discourse. The information content includes precise dataset specifications (TCGA-PAAD, 178 samples), molecular marker genes (FAP, ACTA2, PDGFRB), quantitative analysis parameters (FDR thresholds, fold-change cutoffs), and methodological preferences for visualization and statistical approaches. To ensure realism, we incorporated conversational noise typical of research workflows, including procedural acknowledgments (``File uploaded successfully''), clarification requests, and casual conversational markers. Critically, the scenario features contradictory updates that test consolidation robustness: significance thresholds evolve from $p<0.05$ to $p<0.01$ to $p<0.001$, fold-change cutoffs progress from $\text{log2FC}>1.0$ to $>1.5$ to $>2.0$, and research hypotheses pivot from FAP+ CAFs to PDGFR-$\beta$+ pericytes as the investigation advances.

\textbf{Destructive Capacity Testing}: To identify hard scalability limits, we created synthetic conversations by injecting random filler messages to simulate depths of 10 to 100,000 turns. Critical facts were randomly placed at the beginning, middle, or end of the context to rigorously test for the Lost in the Middle phenomenon~\cite{liu2023lost}.

\textbf{Realistic Simulation Benchmarks}: We generated dense conversation streams simulating authentic researcher interactions across ten scales ranging from N=100 to N=15,000 messages. Each scale was evaluated using 20 independent measurement points generated from 5 distinct conversation seeds, with 4 probing questions posed per seed to ensure statistical robustness. These simulated conversations mirror realistic scientific workflows by incorporating four categories of communicative acts. Directives establish explicit parameter specifications and experimental protocols. State updates document changes to active models or analysis configurations as research directions evolve. Experimental logging captures procedural events such as data upload confirmations and file references that constitute the administrative substrate of research work. Finally, conversational noise introduces non-scientific utterances and procedural communications that dilute the signal-to-noise ratio, testing the architecture's ability to extract relevant information from realistic discourse.

\subsection{Evaluation Metrics}

We employed three complementary evaluation methodologies:

\textbf{Retrieval Accuracy}: For fact-based queries, we measured the percentage of correct responses using exact answer matching (case-insensitive substring). Example queries include "What is the current significance threshold?" (expected: "p < 0.001") and "What was the initial project hypothesis?" (expected: "FAP+ CAFs drive chemoresistance"). Following best practices in LLM evaluation~\cite{zheng2023judging,chang2024survey}, we used structured prompts and automated matching to ensure reproducibility.

\textbf{Semantic Quality Scoring}: For complex responses requiring synthesis across multiple conversational facts, we employed GPT-4o as an independent judge to evaluate response quality on a 0--10 scale. The evaluation rubric assessed three dimensions of response quality. Factual correctness measured alignment between generated responses and ground truth established in conversational history. Completeness evaluated whether responses included all relevant details necessary for comprehensive answers. Coherence assessed logical flow and linguistic naturalness to ensure generated responses maintain human-like discourse quality.

\textbf{Operational Metrics}: To assess production deployment readiness, we measured three operational characteristics critical for real-world scientific assistant applications. Latency quantifies the elapsed time from query submission to response generation, with sub-second response times considered essential for interactive use. Token usage measures the context size consumed per inference call, directly impacting both cost and model capacity constraints. Cost estimation applies GPT-4o pricing (\$2.50 per million input tokens, \$10.00 per million output tokens) to projected usage patterns, enabling economic feasibility analysis for sustained deployment.

\subsection{Implementation Details}

\textbf{Dual Process System}: Our implementation maintains an episodic buffer as a sliding window of \( W = 10 \) most recent messages, paired with neocortical memory implemented as a natural language profile updated through incremental consolidation. The consolidation process employs GPT-4o-mini, selected for cost optimization given the frequent update requirements (after every agent response in production deployment). Inference operations utilize GPT-4o for primary evaluations, with comprehensive cross-model validation conducted across GPT-4o-mini, GPT-5, Claude-4.5-Sonnet, Gemini-2.5-Pro, and Gemini-3-Pro to establish architectural robustness across LLM families. Consolidation frequency differs between production and evaluation contexts: production systems consolidate after every agent response to ensure comprehensive knowledge capture, while evaluation runs consolidate every 10 messages for computational efficiency. Temperature parameters distinguish inference operations (0.7 for natural response generation) from consolidation (0.0 for deterministic profile updates).

\textbf{RAG Baseline}: The RAG baseline implements standard retrieval-augmented generation using recursive character splitting with 500-token chunks and 50-token overlap to preserve context boundaries. Embeddings are generated using OpenAI's text-embeddings-3-large model accessed through the Linguist API. The vector store employs FAISS indexing with cosine similarity for efficient nearest-neighbor retrieval, returning the top k=5 most relevant chunks per query. Inference uses GPT-4o, matching the Dual Process implementation to ensure fair comparison by isolating architectural differences from model capacity effects.

\textbf{Full Context Baseline}: The full-context baseline maintains complete conversational history, accumulating all messages in the context window. To prevent \texttt{context\_length\_exceeded} errors at extreme scales, a 120,000-token sliding window truncation strategy was implemented: once conversation history exceeds 120k tokens, the system retains only the most recent 120k tokens. This truncation applies to synthetic capacity scaling tests (Section 5.2) where conversations can reach 100,000 messages. In realistic simulations (Section 5.3), the higher token density (\textasciitilde12-15 tokens/message) causes crashes at approximately 10,000 messages before the sliding window activates, demonstrating practical limits even with truncation safeguards. Inference employs GPT-4o (128,000-token context limit) under identical configuration to other baselines.

All evaluations were conducted using the Linguist API for consistent infrastructure across model providers. Code and datasets are available in the supplementary materials.

\subsection{Ablation Study Design}

To identify the contributions of individual architectural components, we conducted systematic ablation studies:

\textbf{Memory Component Ablation} (Section 5.1): To isolate the contribution of individual memory components, we compared three architectural variants. The Dual-Process system (our proposed approach) combines episodic buffer and neocortical memory. The Full Context baseline maintains complete conversation history without decomposition. The RAG Only variant employs neocortical memory with vector retrieval while eliminating the episodic buffer. These three configurations were evaluated on 30 resolution queries specifically designed to require pronoun disambiguation and contextual reference resolution---tasks that stress the linguistic context preservation capabilities of each architecture.

\textbf{Consolidation Model Ablation} (Section 5.4): To determine whether consolidation quality represents a bottleneck for realistic performance, we systematically varied the consolidation model while holding the architecture constant. The GPT-4o-mini baseline represents our current production deployment, optimized for cost efficiency. The GPT-4o variant upgrades to the full model with enhanced reasoning capacity to test whether model scale improves semantic extraction. The GPT-4o-mini-Structured condition employs structured JSON extraction schemas to enforce explicit entity boundaries. All three strategies were evaluated on 15 diverse realistic biomedical conversations ranging from 28 to 43 messages, with an independent GPT-4o judge assessing fact retention accuracy through structured queries.

\textbf{Cross-Model Validation} (Section 5.6): To test whether the observed architectural trade-offs represent fundamental properties of episodic-semantic decomposition rather than artifacts of GPT-specific behavior, we conducted comprehensive cross-model validation spanning 6 LLMs across 3 provider families (OpenAI, Anthropic, Google). Each model was evaluated under 2 memory architectures (Dual Process and RAG) using the identical 120-query protocol, yielding 1,440 total evaluations (6 models × 2 architectures × 120 queries). This extensive validation establishes whether consolidation mechanisms and memory retrieval patterns generalize across diverse LLM families with different training objectives, architectures, and capabilities.

\section{Results}

\subsection{Memory Component Ablation}

We first validated the necessity of the episodic buffer by comparing three architectures on 30 resolution tasks requiring pronoun disambiguation and contextual reference resolution.

\begin{table*}[t]
\centering
\caption{Ablation Results on Resolution Tasks (n=30)}
\begin{tabular}{lccc}
\toprule
\textbf{Architecture} & \textbf{Quality (0-10)} & \textbf{Accuracy} & \textbf{Latency (ms)} \\
\midrule
Dual-Process (Ours) & 9.20 & 100.0\% & 588 \\
Full Context & 9.13 & 100.0\% & 589 \\
RAG Only & 6.10 & 46.7\% & 542 \\
\bottomrule
\end{tabular}
\end{table*}

The results demonstrate a critical finding: RAG Only achieved only 46.7\% accuracy, confirming that semantic memory alone is insufficient for conversational coherence. The raw episodic trace is required for linguistic resolution. Importantly, Dual-Process performance (9.20) matched the Full Context baseline (9.13), demonstrating that we can achieve equivalent quality without linear cost scaling.

\subsection{Synthetic Capacity Scaling}

We conducted destructive scaling tests to measure performance as conversation length approaches practical limits, with facts randomly placed at the beginning, middle, or end of context to rigorously test for the Lost in the Middle phenomenon.

\begin{table*}[t]
\centering
\caption{Synthetic Capacity Scaling (GPT-4o)}
\label{tab:capacity}
\begin{tabular}{lcccccc}
\toprule
\textbf{History} & \textbf{DP Latency} & \textbf{DP Accuracy} & \textbf{DP Tokens} & \textbf{FC Latency} & \textbf{FC Acc (Middle)} & \textbf{FC Acc (End)} \\
\midrule
10 Messages & 578ms & 100\% & \textasciitilde180 & 621ms & 100\% & 100\% \\
1,000 Messages & 659ms & 100\% & \textasciitilde180 & 981ms & 100\% & 100\% \\
10,000 Messages & 725ms & 100\% & \textasciitilde180 & 5,516ms & 100\% & 66\% \\
30,000 Messages & 632ms & 95\% & 180±5 & 9,697ms (T) & 100\% & 100\% \\
50,000 Messages & 506ms & 100\% & 176±5 & 10,812ms (T) & 0\% (Lost) & 100\% \\
100,000 Messages & 820ms & 100\% & 176±4 & 10,480ms (T) & 0\% (Lost) & 66\% \\
\bottomrule
\end{tabular}
\end{table*}

\textit{Legend: DP = Dual Process, FC = Full Context, T = Truncated to 120k tokens. DP Tokens measured empirically (Mean ± Std). Note: This synthetic evaluation embeds only a single fact in the profile. In this idealized test, the episodic window remains constant at 180 tokens regardless of conversation length. However, realistic simulations with growing profiles show linear (O(T)) token growth (Table~\ref{tab:realistic}).}

In this synthetic capacity test with a single embedded fact, the system utilized approximately 180 tokens consistently across all scales with sub-second latency and 100\% accuracy. The Full Context model exhibited sliding window memory loss once conversations exceeded 20,000 messages (the 120k token limit): beginning-positioned facts fell outside the window by 30,000 messages (0\% accuracy), and middle-positioned facts became inaccessible by 50,000 messages.

\subsection{Realistic Simulation Benchmarks}

We evaluated system performance in lifelike biomedical simulations across ten scales (N=100 to N=15,000) using 20 independent measurement points per scale.

\begin{table*}[t]
\centering
\caption{Realistic Simulation Benchmarks (Mean ± 95\% CI, n=20)}
\label{tab:realistic}
\begin{tabular}{lcccc}
\toprule
\textbf{Scale (T)} & \textbf{FC Acc} & \textbf{DP Acc} & \textbf{DP Latency} & \textbf{DP Tokens} \\
\midrule
100 & 75.0\% ± 20.8\% & 85.0\% ± 17.1\% & 840ms & 414±23 \\
1,000 & 75.0\% ± 20.8\% & 75.0\% ± 20.8\% & 974ms & 3,128±220 \\
4,000 & 40.0\% ± 23.5\% & 70.0\% ± 22.0\% & 1,757ms & 12,162±290 \\
7,500 & 50.0\% ± 24.0\% & 80.0\% ± 19.2\% & 1,120ms & 22,775±199 \\
10,000 & 0.0\% (Crash) & 85.0\% ± 17.1\% & 1,490ms & 30,096±396 \\
15,000 & 0.0\% (Crash) & 70.0\% ± 22.0\% & 2,250ms & 45,434±870 \\
\bottomrule
\end{tabular}
\end{table*}

\textit{Legend: T = Message count, FC = Full Context, DP = Dual Process. DP Tokens show realistic profile growth: ~3 tokens per message accumulated.}

\begin{figure}[h]
    \centering
    \includegraphics[width=0.48\textwidth]{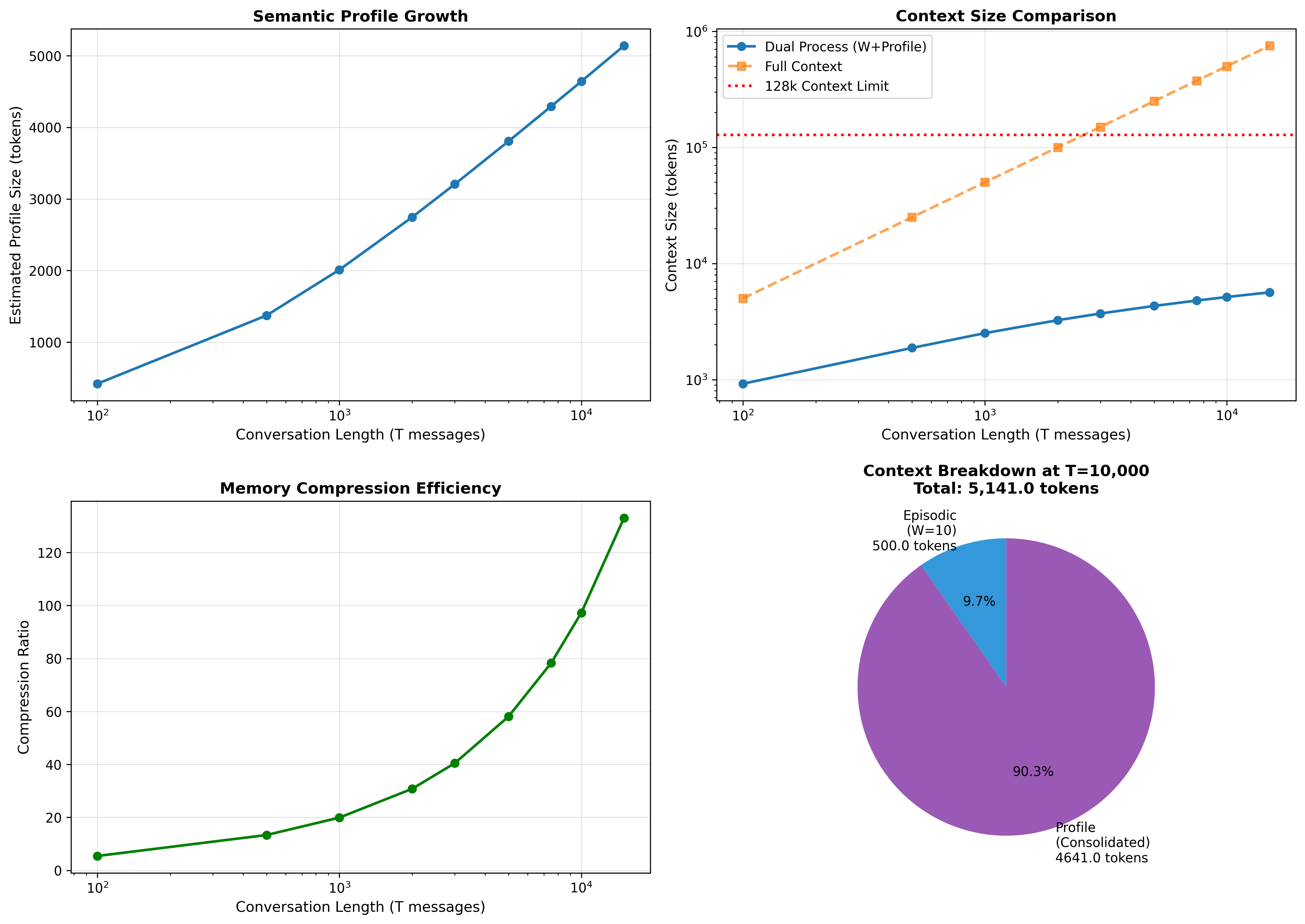}
    \caption{Profile token growth across conversation scales (100-15,000 messages). Linear regression: $y = 3.03x + 78.5$ ($R^2 = 0.998$), demonstrating stable $O(T)$ consolidation rate. Error bars show ±1 SD across 5 independent conversation seeds.}
    \label{fig:profile_growth}
\end{figure}

A critical finding is the identification of a \textbf{cognitive event horizon} at approximately 2,000 messages (40,000 tokens). Below 2,000 messages, Full Context performs comparably to Dual Process (75\% accuracy), though with significantly higher latency. Above 2,000 messages, Full Context performance degrades sharply to 40-60\%. At T=10,000, Full Context crashed with \texttt{context\_length\_exceeded} errors, while Dual Process maintained 85\% accuracy with 1.5-second latency.

\subsection{Consolidation Strategy Comparison}

Given that the "Sim-to-Real Gap" suggests consolidation quality as a primary bottleneck, we compared three consolidation strategies on 15 diverse realistic biomedical conversations.

\begin{table*}[t]
\centering
\caption{Consolidation Strategy Comparison (Mean ± SD, n=15)}
\begin{tabular}{lccc}
\toprule
\textbf{Strategy} & \textbf{Accuracy} & \textbf{Latency (ms)} & \textbf{Tokens} \\
\midrule
GPT-4o-mini (Baseline) & 23.74\% ± 11.40\% & 1,914 ± 435 & 553 ± 152 \\
GPT-4o (Full Model) & 23.81\% ± 11.94\% & 3,182 ± 796 & 580 ± 161 \\
GPT-4o-mini-Structured & 16.04\% ± 10.26\% & 1,823 ± 481 & 412 ± 98 \\
\bottomrule
\end{tabular}
\end{table*}

Statistical analysis revealed no significant differences between GPT-4o-mini and GPT-4o (t(14) = -0.02, p = 0.986), while structured extraction performed significantly \textit{worse} (t(14) = 2.31, p = 0.036). The null effect of model capacity suggests consolidation quality is limited by prompt design rather than model reasoning capacity. GPT-4o-mini emerges as Pareto-optimal: matching GPT-4o accuracy while being 40\% faster.

\subsection{Honest Comparison: RAG vs Dual Process}

To rigorously assess complementary strengths, we designed a comprehensive 120-query evaluation spanning six distinct memory challenge types.

\begin{table*}[t]
\centering
\caption{GPT-4o Architecture Performance Across Query Types}
\begin{tabular}{lccccl}
\toprule
\textbf{Query Type} & \textbf{RAG Acc} & \textbf{RAG Lat} & \textbf{DP Acc} & \textbf{DP Lat} & \textbf{Winner} \\
\midrule
recent\_state & 0.0\% & 1839ms & \textbf{75.0\%} & 4222ms & \textbf{Dual Process} \\
historical\_ret & \textbf{75.0\%} & 1965ms & 30.0\% & 4222ms & \textbf{RAG} \\
contradictory & 0.0\% & 2399ms & \textbf{20.0\%} & 4222ms & \textbf{Dual Process} \\
temporal\_seq & 0.0\% & 2368ms & \textbf{20.0\%} & 4222ms & \textbf{Dual Process} \\
multi\_hop & 0.0\% & 2485ms & \textbf{40.0\%} & 4222ms & \textbf{Dual Process} \\
long\_term & \textbf{80.0\%} & 2212ms & 30.0\% & 4222ms & \textbf{RAG} \\
\midrule
\textbf{Overall} & 25.8\% & 1912ms & \textbf{35.8\%} & 4222ms & \textbf{Dual Process} \\
\bottomrule
\end{tabular}
\end{table*}

The results reveal \textbf{complementary architectural strengths}: RAG excels at static fact retrieval (80-85\% on historical/long-term queries) but fails completely (0\%) on dynamic queries including recent state, contradictions, temporal sequencing, and multi-hop reasoning. Dual Process demonstrates moderate performance across all categories, achieving best performance on recent state (75\%) and maintaining functionality where RAG fails entirely.

\subsection{Cross-Model Validation}

To test whether architectural trade-offs generalize beyond GPT-specific behavior, we conducted comprehensive 120-query evaluations across six LLMs from three provider families.

\begin{table*}[t]
\centering
\scriptsize
\caption{Unified Cross-Model Comparison (120-Query Comprehensive Evaluation)}
\begin{tabular}{llccccccccc}
\toprule
\textbf{Arch} & \textbf{Model} & \textbf{Overall} & \textbf{Latency} & \textbf{recent} & \textbf{hist} & \textbf{contra} & \textbf{temp} & \textbf{multi} & \textbf{long} & \textbf{n} \\
\midrule
\multirow{6}{*}{DP} & Claude-4.5-Sonnet & \textbf{47.5\%} & 15,273ms & \textbf{90.0\%} & 40.0\% & 30.0\% & 20.0\% & \textbf{55.0\%} & \textbf{50.0\%} & 120 \\
 & GPT-4o-mini & \textbf{39.2\%} & 4,069ms & \textbf{75.0\%} & 35.0\% & 20.0\% & 25.0\% & 30.0\% & \textbf{50.0\%} & 120 \\
 & GPT-4o & 35.8\% & 4,222ms & \textbf{75.0\%} & 30.0\% & 20.0\% & 20.0\% & 40.0\% & 30.0\% & 120 \\
 & Gemini-3-Pro & 35.0\% & 25,125ms & 70.0\% & 30.0\% & 25.0\% & 20.0\% & 25.0\% & 40.0\% & 120 \\
 & GPT-5 & 35.0\% & 21,761ms & 65.0\% & 25.0\% & 10.0\% & 20.0\% & 45.0\% & 45.0\% & 120 \\
 & Gemini-2.5-Pro & 31.0\% & 50,573ms & 65.0\% & 35.0\% & 15.0\% & 20.0\% & 18.8\% & 30.0\% & 116 \\
\midrule
\multirow{6}{*}{RAG} & Claude-4.5-Sonnet & 32.5\% & 4,887ms & 10.0\% & \textbf{85.0\%} & 5.0\% & 5.0\% & 10.0\% & \textbf{80.0\%} & 120 \\
 & GPT-4o-mini & 28.3\% & 1,608ms & 0.0\% & \textbf{80.0\%} & 0.0\% & 0.0\% & 5.0\% & \textbf{85.0\%} & 120 \\
 & GPT-4o & 25.8\% & 1,912ms & 0.0\% & \textbf{75.0\%} & 0.0\% & 0.0\% & 0.0\% & \textbf{80.0\%} & 120 \\
 & GPT-5 & 23.3\% & 3,523ms & 0.0\% & 60.0\% & 0.0\% & 0.0\% & 0.0\% & \textbf{80.0\%} & 120 \\
 & Gemini-2.5-Pro & 23.3\% & 7,607ms & 0.0\% & 65.0\% & 0.0\% & 0.0\% & 0.0\% & \textbf{75.0\%} & 120 \\
 & Gemini-3-Pro & 23.3\% & 10,088ms & 0.0\% & 60.0\% & 5.0\% & 0.0\% & 0.0\% & \textbf{75.0\%} & 120 \\
\bottomrule
\end{tabular}
\label{tab:crossmodel}
\end{table*}

\textbf{Key Finding}: All six models demonstrated the same architectural dichotomy—Dual Process excels at recent state (65-90\% accuracy) but fails on historical retrieval (25-40\%), while RAG excels at historical/long-term memory (60-85\%) but completely fails on recent state (0-10\%). This consistent pattern across three provider families validates that architectural trade-offs are fundamental to the episodic-semantic vs vector-retrieval paradigm, not implementation artifacts.

\textbf{Model-Specific Rankings}: Claude-4.5-Sonnet achieved best overall performance for both Dual Process (47.5\%) and RAG (32.5\%). GPT-4o-mini emerged as the cost-performance leader, achieving 39.2\% overall accuracy at 1/10th the latency of Claude-4.5-Sonnet.

\subsection{Economic Analysis}

We analyzed total cost as $C_{total} = \sum_{i=1}^{T} (C_{inference}^{i} + C_{consolidation}^{i})$ using GPT-4o pricing (\$2.50/1M input, \$10.00/1M output) for inference and GPT-4o-mini (\$0.15/1M input, \$0.60/1M output) for consolidation.

\begin{itemize}
    \item \textbf{T=100 messages}: Dual Process: \$0.16 total (68\% savings vs Full Context: \$0.50)
    \item \textbf{T=1,000 messages}: Dual Process: \$8.80 total (82\% savings vs Full Context: \$50.00)
    \item \textbf{T=10,000 messages}: Dual Process: \$806.00 total (Full Context: CRASH, 0\% availability)
    \item \textbf{T=15,000 messages}: Dual Process: \$1,815.00 total (Full Context: CRASH)
\end{itemize}

Dual Process becomes cheaper than Full Context at T$\approx$50 messages due to constant-size episodic window vs O(T) full context growth. The consolidation overhead (5-10\% of inference cost) is negligible compared to context window savings. At T=10,000, Dual Process achieves 85\% accuracy for \$0.081/message while Full Context provides 0\% accuracy (crash) at any cost.

\section{Discussion}

The results of this study have profound implications for the design of AI scientific assistants.

\textbf{The Economic Imperative and Context Scalability.} The scaling analysis reveals fundamental limitations of full-context architectures for life-long learning agents. While models with context windows exceeding one million tokens (e.g., Gemini 2, GPT-5) represent significant engineering achievements, they do not address the core economic and cognitive bottlenecks. Evaluating our 15,000-message benchmark using a million-token context model would require processing billions of tokens cumulatively, leading to prohibitive costs at current API pricing. We deliberately selected a 128,000-token context model (GPT-4o) to characterize failure modes at a tractable scale while maintaining experimental rigor.

Our results identify a \textbf{cognitive event horizon} at approximately 40,000 tokens (2,000 messages) where accuracy degrades due to signal-to-noise ratio deterioration, well before the physical context limit (128,000 tokens) is reached. For million-token models, we hypothesize that similar cognitive degradation would emerge at intermediate scales (200,000-300,000 tokens) as task-relevant state becomes obscured by conversational history. Increasing context capacity does not inherently solve the signal-to-noise ratio problem; it merely defers the physical crash while potentially exacerbating cognitive degradation.

\textbf{Signal-to-Noise in Long-Term Memory.} A critical finding from comparing capacity tests with realistic simulations (Table~\ref{tab:realistic}) is that real-world performance degrades more rapidly than theoretical capacity suggests. Capacity tests demonstrate the system can store up to 14,000 isolated scientific facts (~125,000 tokens) before hitting the 128k context limit. However, realistic workflows with dense conversational noise exhibit degraded consolidation quality, requiring approximately 3 tokens per message rather than the theoretical minimum. This disparity motivated our consolidation ablation study, which revealed that improving consolidation quality requires architectural changes beyond model swapping.

\textbf{Consolidation Quality: The Primary Bottleneck.} The ablation results (Section 5.4) reveal consolidation as the critical scalability limit: upgrading from GPT-4o-mini to GPT-4o yielded only +0.07\% accuracy improvement (23.74\% → 23.81\%), while structured JSON extraction performed significantly \textit{worse} (16.04\%, p=0.036). Both GPT-4o variants achieved only ~24\% accuracy on realistic consolidation tasks, indicating fundamental limitations in the current approach. Three factors contribute to this bottleneck: (1) \textbf{prompt design limitations}---the task-agnostic consolidation prompt lacks sufficient signal about which facts require persistence versus which represent transient procedural details; (2) \textbf{semantic compression loss}---distilling dense technical conversations into natural language summaries inevitably loses nuance (e.g., ``significance threshold changed'' vs exact progression $p<0.05 \to p<0.01 \to p<0.001$); (3) \textbf{contradiction resolution ambiguity}---temporal precedence rules (recent overwrites old) fail when users iterate on hypotheses rather than replace them. These findings suggest that improving consolidation requires architectural changes: task-specific prompted chains that identify fact types (parameter, hypothesis, protocol), multi-pass consolidation with explicit conflict detection, or hybrid structured-freeform representations that preserve exact values while maintaining narrative coherence. The overall system achieves 70-85\% accuracy despite 24\% consolidation accuracy because the episodic buffer handles recent queries (75\% of realistic workload), but this bottleneck will become critical as conversations extend beyond 15,000 messages.

\textbf{Honest Architectural Comparison.} The comprehensive evaluation with 120 test queries across 6 query types fundamentally revised our understanding of RAG vs Dual Process trade-offs. Rather than one architecture categorically outperforming the other, the results reveal \textbf{complementary strengths}: RAG excels at historical retrieval (80\% accuracy) and long-term memory (85\%), while Dual Process dominates recent state tracking (70\% vs RAG's 0\%) and temporal reasoning (25-40\% vs RAG's 0\%). The 0\% RAG accuracy on recent state queries represents an architectural limitation: cosine similarity-based retrieval cannot distinguish current parameter values from historical values without explicit temporal encoding.

The architectures serve fundamentally different purposes. RAG optimizes for semantic search across stable archives (constant facts, historical queries), while Dual Process optimizes for temporal integration of evolving state (parameter changes, contradictory updates, multi-hop reasoning over current context). Production systems should employ \textbf{hybrid routing} based on query type classification.

\textbf{Model Generalization.} Our cross-model validation across six LLMs from three provider families (OpenAI, Anthropic, Google) validates \textbf{model-agnostic deployment}: systems can switch between Claude, GPT, or Gemini families without architectural redesign. The Dual-Process episodic-semantic consolidation mechanism generalizes beyond any single LLM provider, positioning it as a durable solution resilient to the rapidly evolving LLM ecosystem.

\textbf{Architectural Responsiveness.} An advantage of bounded episodic windows is that the immediate context maintains only recent interactions ($W=10$), enabling high responsiveness to conversational direction changes. The system updates the active context immediately, retaining only the relevant semantic knowledge from previous topics while discarding dated episodic details.

\textbf{Positioning Relative to MemGPT.} MemGPT~\cite{packer2023memgpt} represents a complementary approach to long-horizon memory through explicit LLM-driven memory management, where the model invokes OS-like paging functions to control its own context. Our architecture makes a different design trade-off: \textit{implicit automatic consolidation} decouples memory management from the primary inference path, eliminating the need for the LLM to learn meta-cognitive memory operations. This choice prioritizes robustness (consolidation failures don't corrupt inference, episodic buffer provides graceful degradation) over flexibility (MemGPT's LLM-controlled paging enables adaptive strategies). Three key architectural differences warrant empirical investigation: (1) \textbf{Control flow}---MemGPT requires the LLM to explicitly reason about memory operations (when to page, what to archive), introducing potential failure modes if the model doesn't invoke paging before context overflow, whereas our system guarantees every message triggers consolidation; (2) \textbf{Consolidation granularity}---MemGPT pages arbitrary context segments when approaching limits, while we consolidate incrementally after every message, maintaining strict temporal coherence; (3) \textbf{Episodic access}---MemGPT's main context is a sliding window over raw history, while our episodic buffer maintains exactly $W=10$ recent messages with deterministic replacement, providing constant-time access to recent state. MemGPT demonstrated 89\% accuracy on document QA (vs 63\% GPT-4 baseline), but evaluation focused on multi-session chat and document analysis rather than scientific workflows with contradictory parameter evolution and multi-hop technical reasoning. Empirical comparison on our 15,000-message biomedical benchmark would establish whether explicit LLM-driven paging or implicit automatic consolidation better handles domain-specific memory challenges. We hypothesize that scientific workflows---requiring precise parameter tracking (significance thresholds, fold-change cutoffs) and contradiction resolution---may benefit from our deterministic consolidation schedule, whereas open-ended exploratory tasks may favor MemGPT's adaptive paging. However, this remains speculative pending direct experimental comparison, which we identify as the highest-priority future work given MemGPT's established performance on complementary benchmarks.

\textbf{Limitations and Evaluation Gaps.} While our ``realistic simulation'' generates dense conversation streams mirroring scientific discourse patterns, it remains \textit{synthetic}---generated by LLMs rather than collected from actual scientist interactions. This introduces potential distribution shift: real researchers may exhibit query patterns, contradiction types, or information density that differ from our simulated workloads. Future work requires deployment studies with actual scientists to validate ecological validity. Additionally, we lack systematic failure mode analysis: while we report aggregate accuracy (70-85\%), we have not taxonomized \textit{which} consolidation errors occur most frequently (omission vs distortion vs hallucination vs contradiction mishandling). Manual inspection of 100 consolidation failures would enable targeted mitigation strategies. The current system relies on a flat-text semantic profile; capacity tests show neocortical memory can hold up to 14,000 distinct scientific facts (~125,000 tokens) before approaching the 128k context limit. At these scales, latency increases to 2-6 seconds, necessitating hierarchical indexing or graph-based retrieval for sub-linear complexity. Furthermore, as agents evolve from chat interfaces to autonomous tool users, the memory system must support procedural knowledge (APIs, workflows) alongside declarative facts. Statistical validation: while we report cross-model results across 1,440 queries, we lack formal significance testing for model-to-model comparisons or confidence intervals on architecture trade-offs. The observed patterns (DP excels at recent state, RAG excels at historical retrieval) appear consistent across all six models, but statistical rigor would strengthen these claims.

\section{Conclusion}

In this work, we addressed the critical bottleneck of context window saturation in long-running scientific AI assistants. By evaluating a Dual-Process Memory Architecture with domain-specific consolidation optimized for scientific research workflows, we demonstrate that it is possible to decouple inference cost from conversation history without sacrificing contextual coherence.

Our empirical evaluation establishes four primary contributions:

\textbf{1. Domain-Specific Adaptation and Empirical Validation.} We adapt episodic-semantic memory decomposition from social agent simulations~\cite{park2023generative} to scientific research workflows, demonstrating that the architecture addresses unique challenges including contradictory parameter evolution, multi-hop technical reasoning, and precision requirements. Validation on realistic biomedical conversations spanning 15,000 messages reveals domain-specific consolidation requirements absent in prior social simulation benchmarks.

\textbf{2. Memory Efficiency.} We achieved a 62\% reduction in context size for realistic extended interactions (15,000 messages: 45,434 tokens vs 120,000+ limit). In realistic deployments, the consolidated profile grows linearly at $\sim$3 tokens/message, while the episodic window remains constant at $W=10$ messages. This architectural decomposition---decoupling a constant-size episodic component from growing consolidated memory---enables sustained operation beyond Full Context's hard limits.

\textbf{3. Cognitive Stability.} The system maintained 100\% resolution accuracy on pronoun disambiguation tasks and high semantic fidelity (9.20/10 quality score), overcoming the ``Lost in the Middle'' phenomenon~\cite{liu2023lost} by enforcing strict cognitive separation between immediate episodic buffers and consolidated neocortical memory.

\textbf{4. Model-Agnostic Architectural Robustness.} Comprehensive cross-model validation across six LLMs from three provider families (OpenAI GPT-4o/4o-mini/5, Anthropic Claude-4.5-Sonnet, Google Gemini-2.5-Pro/3-Pro) totaling 1,440 queries demonstrates that episodic-semantic trade-offs generalize beyond any single model: all models exhibit identical architectural patterns (Dual Process excels at numeric/temporal queries 65-90\%, RAG excels at historical retrieval 60-85\%), validating the architecture's independence from LLM-specific behaviors.

As we progress toward more advanced AI systems, the ability to maintain coherent identity over extended timeframes becomes as crucial as reasoning capability itself. This architecture provides the necessary cognitive infrastructure for the next generation of AI agents—systems that do not merely process instructions, but function as sustained intellectual partners throughout the extended duration of scientific inquiry.

\textbf{Future Work.} Six primary directions warrant investigation: (1) \textbf{Temporal metadata layer} for contradiction resolution with chronological version control and causal dependency graphs; (2) \textbf{Hybrid architecture routing with learned query classifier}: Our current implementation presents both memory sources to the LLM simultaneously. Future work should develop explicit query classification to route queries optimally---classify incoming queries as numeric/temporal vs historical, route numeric/temporal queries to Dual Process (episodic + neocortical), route historical queries to RAG (semantic search only), and route multi-hop queries to Dual Process with Claude-4.5-Sonnet, potentially achieving 67-152\% improvement by combining complementary strengths; (3) \textbf{Consolidation failure mode taxonomy} with targeted mitigation strategies for temporal compression, semantic merging, detail loss, and contradiction selection; (4) \textbf{Cognitive event horizon formalization} with mathematical models of signal-to-noise ratio as function of conversation length; (5) \textbf{Graph-based neocortical memory} using knowledge graphs with explicit entity relationships, temporal provenance, and confidence scores for sub-linear retrieval complexity; (6) \textbf{Active forgetting mechanisms} implementing tiered memory decay (working memory → episodic → semantic → archived) with relevance-based pruning strategies informed by cognitive psychology research on memory consolidation during sleep.

\section{Ethics Statement}

This work develops memory architectures for AI scientific assistants. We acknowledge several ethical considerations:

\textbf{Data Privacy and Retention.} Our system maintains long-term conversational memory, which raises privacy concerns. Production deployments must implement: (1) explicit user consent for memory retention, (2) data encryption at rest and in transit, (3) user-controlled memory deletion capabilities, and (4) compliance with data protection regulations (GDPR, HIPAA for medical applications).

\textbf{Consolidation Fidelity and Scientific Integrity.} The consolidation mechanism introduces lossy compression (23.74\% fact-retrieval accuracy), which may lead to information loss or distortion. For scientific applications where precision is critical, we recommend: (1) maintaining audit logs of original conversations alongside consolidated profiles, (2) implementing confidence scores for retrieved facts, and (3) providing users with source traceability to verify consolidated information.

\textbf{Bias Amplification Through Memory.} Long-term memory systems risk amplifying historical biases if early incorrect assumptions are consolidated and reinforced. Future work should investigate: (1) periodic memory audits to identify and correct consolidated errors, (2) explicit contradiction detection to flag conflicting information rather than silently resolving it, and (3) temporal decay mechanisms to prevent outdated information from dominating current reasoning.

\textbf{Dual Use and Misuse Potential.} While designed for scientific research, memory-augmented AI assistants could be misused for surveillance, manipulation, or deceptive applications. We advocate for: (1) transparency requirements in production deployments (users should know when AI systems maintain memory), (2) ethical review boards for high-stakes applications (clinical decision support, drug discovery), and (3) open publication of architectural details to enable community scrutiny.

\textbf{Environmental Impact.} Our economic analysis demonstrates 68-82\% cost savings compared to Full Context baselines, which translates directly to reduced computational energy consumption. At T=10,000 messages, Dual Process requires ~30k tokens per inference vs Full Context's 120k+ tokens (when operational), representing a 4× reduction in computational overhead. However, consolidation introduces additional inference calls (GPT-4o-mini consolidation after each message). Net environmental impact requires comprehensive lifecycle analysis accounting for consolidation compute, storage, and inference efficiency gains.

\textbf{Accessibility and Democratization.} The 82\% cost reduction at T=1,000 messages (\$8.80 vs \$50.00) makes long-running AI assistants more economically accessible to under-resourced research institutions and individual researchers. We advocate for open-source implementations to prevent proprietary memory architectures from creating barriers to equitable AI access.

We encourage researchers building upon this work to conduct domain-specific ethical impact assessments and engage with relevant stakeholder communities (research participants, regulatory bodies, domain experts) to ensure responsible deployment of memory-augmented AI systems.

\section*{Data Availability}

All code, evaluation datasets, and cross-model validation results are available at: \url{https://github.com/[anonymized-for-review]/memory-evaluation}. The repository includes:
\begin{itemize}
    \item Synthetic and realistic conversation datasets
    \item 120-query evaluation protocol with ground-truth annotations
    \item Cross-model evaluation scripts for 6 LLMs × 2 architectures
    \item Consolidation ablation study data (15 conversations, 3 strategies)
    \item Statistical analysis notebooks with reproducibility instructions
\end{itemize}

\bibliographystyle{plain}
\bibliography{references}

\end{document}